\documentclass{bmvc2k}

\title{How Important is Importance Sampling for Deep Budgeted Training?}

\addauthor{Eric Arazo}{eric.arazo@insight-centre.org}{1}
\addauthor{Diego Ortego}{diego.ortego@insight-centre.org}{1}
\addauthor{Paul Albert}{paul.albert@insight-centre.org}{1}
\addauthor{Noel E. O'Connor}{noel.oconnor@insight-centre.org}{1}
\addauthor{Kevin McGuinness}{kevin.mcguinness@insight-centre.org}{1}

\addinstitution{
 School of Electronic Engineering,\\
 Insight SFI Centre for Data Analytics,\\
 Dublin City University (DCU),\\
 Dublin, Ireland
}

\runninghead{Arazo, Ortego, Albert, O'Connor, McGuinness}{Importance Sampling}

\usepackage{microtype}
\usepackage{booktabs}
\usepackage{tabularx}  
\usepackage{wrapfig}

\usepackage{soul}
\usepackage{color}

\DeclareRobustCommand{\ctB}[1]{{\color{blue}#1}}

\begin{document}

\maketitle

\begin{abstract}
Long iterative training processes for Deep Neural Networks (DNNs) are commonly required to achieve state-of-the-art performance in many computer vision tasks. Importance sampling approaches might play a key role in budgeted training regimes, i.e.~when limiting the number of training iterations. These approaches aim at dynamically estimating the importance of each sample to focus on the most relevant and speed up convergence. This work explores this paradigm and how a budget constraint interacts with importance sampling approaches and data augmentation techniques. We show that under budget restrictions, importance sampling approaches do not provide a consistent improvement over uniform sampling. We suggest that, given a specific budget, the best course of action is to disregard the importance and introduce adequate data augmentation; e.g.~when reducing the budget to a 30\% in CIFAR-10/100, RICAP data augmentation maintains accuracy, while importance sampling does not. We conclude from our work that DNNs under budget restrictions benefit greatly from variety in the
training set and that finding the right samples to train on is not the most effective strategy when balancing high performance with low computational requirements. Source code available at: \url{https://git.io/JKHa3}
\end{abstract}


\section{Introduction}
The availability of vast amounts of labeled data is crucial when training deep neural networks (DNNs)~\cite{2018_ECCV_limits_weakly, 2020_CVPR_improveImagenet}. Despite prompting considerable advances in many computer vision tasks~\cite{ 2018_ECCV_captioning, 2019_CVPR_pose_Estimation}, this dependence poses two challenges: the generation of the datasets and the large computation requirements that arise as a result. Research addressing the former has experienced great progress in recent years via novel techniques that reduce the strong supervision required to achieve top results~\cite{2019_ICML_efficientnet, 2019_NeurIPS_fixing} by, e.g.~improving semi-supervised learning~\cite{2019_NIPS_MixMatch, 2020_IJCNN_pseudolab}, self-supervised learning~\cite{CVPR_2020_moco, CVPR_2020_InvRepresentations}, or training with noisy web labels~\cite{2019_ICML_BynamicBootstrapping, 2020_ICLR_DivideMix}. The latter challenge has also experienced many advances from the side of network efficiency via DNN compression~\cite{2018_ICML_compressing, 2019_CVPR_compressGAN}, neural architecture search~\cite{2019_ICML_efficientnet, 2019_ICLR_proxylessnas}, or parameter quantization~\cite{2016_ECCV_xnor, 2018_CVPR_quantization}.
All these approaches are designed with a common constraint: a large dataset is needed to achieve top results~\cite{2020_CVPR_improveImagenet}. This conditions the success of the training process on the available computational resources. Conversely, a smart reduction of the amount of samples used during training can alleviate this constraint~\cite{2018_ICML_notAllSamples, 2020_ICML_coreSet}.

The selection of samples plays an important role in the optimization of DNN parameters during training, where Stochastic Gradient Descent (SGD)~\cite{2012_NeurIPS_sgd, 2018_SIAM_optim} is often used. SGD guides the parameter updates using the estimation of model error gradients over sets of samples (mini-batches) that are uniformly randomly selected in an iterative fashion. This strategy assumes equal importance across samples, whereas other works suggest that alternative strategies for revisiting samples are more effective in achieving better performance \cite{2017_NeurIPS_activeBias, 2020_AISTATS_osgd} and faster convergence \cite{2018_ICML_notAllSamples, 2019_Arxiv_selective}. Similarly, the selection of a unique and informative subset of samples (core-set) \cite{2018_ICLR_forget, 2020_ICLR_proxySelection} can reduce the computation requirements during training, while reducing the performance drop with respect to training on all data. However, although removing data samples speeds up training, precise sample selection often requires a pretraining stage that acts counter computational reduction~\cite{2020_ICML_coreSet, 2018_ICLR_activelearning}. 

A possible solution to this limitation might be to dynamically change the important subset during training, as is done by importance sampling methods~\cite{2017_CEMNLP_repeat, 2019_NeurIPS_autoAssist}, which select the samples based on a sampling distribution that evolves with the model and often depends on the loss or network logits~\cite{2015_ICLRw_online, 2018_NeurIPS_RAIS}. An up-to-date sample importance estimation is key for current methods to succeed but, in practice, is infeasible to compute~\cite{2018_ICML_notAllSamples}. The importance of a sample changes after each iteration and estimations become out-dated, yielding considerable performance drops~\cite{2017_NeurIPS_activeBias, 2019_NeurIPS_autoAssist}. Importance sampling methods focus on training with the most relevant samples and achieve a convergence speed-up as a side effect. They do not, however, strictly study the benefits on DNN training when restricting the number of training iterations, i.e.~the budget.

Budgeted training~\cite{2017_NeurIPS_adaptive_budget, 2019_ICML_opportunistic_budget, 2020_ICLR_budget} imposes an additional constraint on the optimization of a DNN: a maximum number of iterations. Defining this budget provides a concise notion of the limited training resources. Li et al.~\cite{2020_ICLR_budget} propose to address the budget limitation using specific learning rate schedules that better suit this scenario. Despite the standardized scenario that budgeted training poses to evaluate methods when reducing the computation requirements, there are few works to date in this direction~\cite{2020_ICLR_budget, 2018_ICML_notAllSamples}. As mentioned, importance sampling methods are closely related, but the lack of exploration of different budget restrictions makes these approaches less applicable: the sensitivity to hyperparamenters that they often exhibit limits their generalization~\cite{2017_NeurIPS_activeBias, 2015_ICLRw_online}.

In this paper, we overcome the limitations outlined above by analyzing the effectiveness of importance sampling methods when a budget restriction is imposed~\cite{2020_ICLR_budget}. Given a budget restriction, we study synergies among importance sampling and data augmentation~\cite{2018_ACML_RICAP, 2020_CVPRw_randAugm, 2018_ICLR_mixup}. We find the improvements of importance sampling approaches over uniform random sampling are not always consistent across budgets and datasets. We argue and experimentally confirm (see Section~\ref{sub_sec:EXP_VAR}) that when using certain data augmentation strategies~\cite{2018_ACML_RICAP, 2020_CVPRw_randAugm, 2018_ICLR_mixup}, existing importance sampling techniques do not provide further benefits, making data augmentation the most effective strategy to exploit a given budget. 

\section{Related work} 
Few works exploit a budgeted training paradigm \cite{2020_ICLR_budget}. Instead, many aim to speed up convergence to a given performance using a better sampling strategy or carefully organizing the samples to allow the model to learn faster and generalize better. Others explore how to improve model performance by labeling the most important samples from an unlabeled set  \cite{2019_CVPR_learning_loss_for_active_learning, 2020_ICLR_deepactivelearning, 2020_ArXiv_active_learning_survey} or how to better train DNNs when limited samples per class are available \cite{2020_CVPR_fewshot_base, 2021_IJCNN_relab, 2019_ICLR_closer_look_to_activeLearning}. None of these works, however, explore the efficiency of these approaches when
trained under constraints in the number of iterations allowed, i.e.~budgeted training. This section reviews relevant works that aim to improve the computational efficiency of training DNNs.

\textbf{Curriculum learning (CL)} aims to improve model performance by ordering the samples from easy to difficult~\cite{2018_ICML_CLTransfLearn, 2009_ICML_CL, 2019_ICML_PowerOfCL, 2019_CVPR_L2G}. Like importance sampling approaches, CL leverages different samples at different stages of training. However, while CL prioritizes easy samples at the beginning of training and includes all of them at the end, importance sampling prioritizes the most difficult subset of samples at each stage of the training. The main drawback of CL is that, in most cases, the order of the samples (curriculum) has to be defined before training, which is already a costly task that requires manually assessing the sample difficulty or transferring knowledge from a pre-trained model. Some approaches remedy this with a simple curriculum \cite{2017_ICCV_focal_loss} or by learning it during training \cite{2018_ICML_mentorNet}; these methods, however, do not aim to speed up training by ordering the samples, but to improve  convergence by weighting the sample contribution to the loss.

\textbf{Core-set selection approaches} aim to find the subset of samples that is most useful~\cite{2018_ICLR_forget, 2020_ICLR_proxySelection, 2020_ICML_coreSet} and maintain accuracy despite training on a fraction of the data. The ability of these methods to reduce training cost relies on using smaller training sets, but the benefit is limited since they require a pre-training stage with the full dataset. They do, however, demonstrate that DNNs can achieve peak performance with a fraction of the full dataset. Some approaches to core-set selection use the most often forgotten samples by the network~\cite{2018_ICLR_forget}, the nearest samples to cluster centroids built from model features~\cite{2020_ICML_coreSet}, or use a smaller pretrained model to select the most informative samples~\cite{2020_ICLR_proxySelection}.

\textbf{Importance sampling} approaches lie in the middle ground between the previous two: they aim to speed up training convergence by leveraging the most useful samples at each training stage~\cite{2018_ICML_notAllSamples, 2019_Arxiv_selective, 2019_NeurIPS_autoAssist},
which correspond to those with highest loss gradient magnitude~\cite{2014_NeurIPS_sgdweight, 2015_ICML_stochasticIS, 2016_ICLR_varianceRed}. Johnson and Guestrin~\cite{2018_NeurIPS_RAIS} have shown that the last layer gradients are a good approximation and are easier to obtain in deep learning frameworks. Alternative importance measures include the loss \cite{2019_Arxiv_selective}, the probability predicted for the true class \cite{2017_NeurIPS_activeBias}, or the rank order of these probabilities~\cite{2015_ICLRw_online}.

The approximation of the optimal distribution by importance sampling approaches avoids the cost of computing the importance of each sample at every iteration. However, this distribution changes very rapidly between iterations, leading to outdated estimations. Initial attempts at addressing this included using several hyper-parameters to smooth the estimated distribution~\cite{2017_NeurIPS_activeBias}, more frequent distribution updates via additional forward passes~\cite{2015_ICLRw_online}, or alternative measures to estimate the sampling distribution~\cite{2017_CEMNLP_repeat}. Several works added complex support techniques to the training to estimate a better distribution: using robust optimization~\cite{2018_NeurIPS_RAIS}, introducing repulsive point techniques~\cite{2019_AAAI_repulsive}, or adding a second network~\cite{2019_NeurIPS_autoAssist}. 

More recent methods leverage the random-then-greedy technique~\cite{2018_Arxiv_randGradBoost}, where the probabilities of an initial random batch of samples are computed and then used to select a batch for training. Within this scheme,~\cite{2018_ICML_notAllSamples} define a theoretical bound for the magnitude of the gradients that allows for faster computation of the sampling probabilities, and~\cite{2019_Arxiv_selective, 2019_ICAAI_biasedSampling} use the loss as a measure of sample importance to keep the sampling distribution updated through the training. Finally, Kawaguchi and Lu~\cite{2020_AISTATS_osgd} introduce the top-$k$ loss~\cite{2017_NeurIPS_topKLoss} to perform the back-propagation step using the samples with highest losses only. Note that these methods do a full forward pass every epoch to update the sampling probabilities.

\textbf{Learning rate schedules} have proven to be useful alternatives for faster convergence. In particular,~\cite{2019_ISOP_superConv, 2017_WACV_cyclical} propose a cyclic learning rate schedule to reach faster convergence by using larger learning rates at intermediate training stages and very low rates at final stages. Similarly, Li et al.~\cite{2020_ICLR_budget} explore budgeted training and propose a linearly decaying learning rate schedule that approaches zero at the end of the training, which without additional hyper-parameters, provides better convergence than the standard learning rate schedulers. These approaches, however, do not explore sample selection techniques to further increase convergence speed.

\section{Budgeted training}\label{sec:BT}
This section formally introduces budgeted training and the different importance sampling methods used through the paper to explore the efficiency of these approaches under budget restrictions. The standard way of training DNNs is by gradient based minimization of cross-entropy
\begin{equation}
\ell(\theta)=-\frac{1}{N}\sum_{i=1}^{N}y_{i}^{T}\log h_{\theta}(y|x_{i}),
\end{equation}
where $N$ is the number of samples in the dataset $D = \left\{ (x_i, y_i) \right\} _{i=1}^{N}$ and $y_i \in \left\{0, 1\right\}^C$ is the one-hot encoding ground-truth label for sample $x_i$, $C$ is the number of classes, $h_{\theta}(y|x_{i})$ is the predicted posterior probability of a DNN model given $x_i$ (i.e.~the prediction after softmax normalization), and $\theta$ are the parameters of the model. Convergence to a reasonable level of performance usually determines the end of the training, whereas in budgeted training there is a fixed iteration budget. We adopt the setting defined by \cite{2020_ICLR_budget}, where the budget is defined as a percentage of the full training setup. Formally, we define the budget $B\in[0,1]$ as the fraction of forward and backward passes used for training the model $h_\theta(x)$ with respect to a standard full training. As we aim at analyzing importance sampling, the budget restriction will be mainly applied to the amount of data $N\times B$ seen every epoch. However, a reduction on the number of epochs $T$ to $T\times B$ (where an epoch $T$ is considered a pass over all samples) is also considered as truncated training for budgeted training.

\textbf{Truncated training} is the simplest approach to budgeted training: keep the standard SGD optimization and reduce the number of epochs trained by the model to $T \times B$. We call this strategy, where the model sees all the samples every epoch, \textit{scan-SGD}. 
While seeing all the samples is common practice, we remove this constraint and draw the samples from a uniform probability distribution at every iteration and call this strategy \textit{unif-SGD}. In this approach the budget is defined by randomly selecting $N \times B$ samples every epoch (and still training for $T$ epochs). 

\textbf{Importance sampling} aims to accelerate the convergence of SGD by sampling the most difficult samples $D_{S} = \left\{(x_i, y_i)\right\}_{i=1}^{N_S}$ more often, where $N_S = N \times B$ (the number of samples selected given a certain budget). Loshchilov and Hutter~\cite{2015_ICLRw_online} proposed a simple approach for importance sampling that uses the loss of every sample as a measure of the sample importance. Chang et al.~\cite{2017_NeurIPS_activeBias} adapts this approach to avoid additional forward passes by using as importance:
\begin{equation}
p_{i}^{t} = \frac{1}{t}\sum_{k=1}^{t} \left(1 - y_{i}^{T}h_{\theta}^{k}(y|x_{i})\right) + \epsilon^{t}, \label{eq1:p-measure}
\end{equation}
where $h_{\theta}^{k}(y|x_{i})$ is the prediction of the model given the sample $x_{i}$ in epoch $k$, and $t$ is the current epoch. Therefore, the average predicted probability across previous epochs associated to the ground-truth class of each sample defines the importance of sample $x_{i}$. The smoothing constant $\epsilon^{t}$ is defined as the mean per sample importance up to the current epoch: $\frac{1}{N}\sum_{i = 1}^{N}p_{i}^{t}$.
The sampling distribution $P^{t}$ at a particular epoch $t$ is then given by:
\begin{equation}
P_{i}^{t} = \frac{p_{i}^{t}}{\sum_{j=1}^{N}p_{j}}. \label{eq3:p-dist}
\end{equation}
By drawing samples from the distribution $P^{t}$ this approach biases the training towards the most difficult samples, and selects samples with highest loss value; we name this method \textit{p-SGD}. Similarly, Chang et al.~\cite{2017_NeurIPS_activeBias} propose to select those samples that are closer to the decision boundaries and favor samples with higher uncertainty by defining the importance measure as $c^{t}_{i} = p^{t}_{i} \times (1-p^{t}_{i})$; we name this approach \textit{c-SGD}.

Both \textit{p-SGD} and \textit{c-SGD} are very computationally efficient as the importance estimation only requires information available during training. Conversely, Jiang et al.~\cite{2019_Arxiv_selective} propose to perform forward passes on all the samples to determine the most important ones and later reduce the amount of backward passes; they name this method selective backpropagation (\textit{SB}). At every forward pass, \textit{SB} stores the sample $x_i$ with probability: 
\begin{equation}
s_{i}^{t} = \left[F_R(\ell(h_{\theta}^{t}(x_{i}), y_i))\right]^b, \label{eq4:SB-prob}
\end{equation}
where $F_R$ is the cumulative distribution function from a history of the 
loss values of the
last $R$ samples seen by the model and $b > 0$ is a constant that determines the selectivity of the method, i.e.~the budget used during the training. In practice, \textit{SB} does as many forward passes as needed until it has enough samples to form a full a mini-batch. It then performs the training forward and backward passes with the selected samples to update the model.

Finally, as an alternative training paradigm to prioritize the most important samples, Kawaguchi and Lu~\cite{2020_AISTATS_osgd} propose to use only the $q$ samples with highest loss from a mini-batch in the backward pass. As the training accuracy increases, $q$ decreases until only $1/16$ of the images in the mini-batch are used in the backward pass. The authors name this approach \textit{ordered} SGD (\textit{OSGD}) and provide a default setting for the adaptive values of $q$. 

\textbf{Importance sampling methods under budgeted training} give a precise notion of the training budget. For \textit{unif-SGD}, \textit{p-SGD}, and \textit{c-SGD} the adaptation needed consists of selecting a fixed number of samples $N\times B$ per epoch based on the corresponding sampling probability distribution $P_{t}$ and still train the full $T$ epochs. For \textit{SB}, the parameter $b$ determines the selectivity of the algorithm: higher values will reject more samples. Note that this method requires additional forward passes that we exclude from the budget as they do not induce the backward passes used for training. By assuming that each DNN backward pass is twice as computationally expensive as a forward pass~\cite{2018_ICML_notAllSamples} we could approximate the budget used by \textit{SB} as $B_{SB} = B + 1/3$, e.g. the results under $B = 0.2$ for \textit{SB} correspond to $B \approx 0.5$ for the other approaches. We adapt \textit{OSGD} by truncating the training as in \textit{scan-SGD}: all the parameters are kept constant but the total number of epochs is reduced to $T \times B$. We also consider the wall-clock time with respect to a full budget training as a metric to evaluate the approaches.


\section{Experiments and Results}
\subsection{Experimental framework}\label{sub_sec:exp_framework}
\textbf{Datasets.~}
We experiment on image classification tasks using CIFAR-10/100~\cite{2009_CIFAR}, SVHN~\cite{2011_NeurIPS_SVHN}, and mini-ImageNet~\cite{2016_NIPS_MiniImageNet} datasets. CIFAR-10/100 consist of 50K samples for training and 10K for testing; each divided into 10(100) classes for CIFAR-10(100). The samples are images extracted from ImageNet \cite{2009_CVPR_ImageNet} and down-sampled to 32$\times$32. SVHN contains 32$\times$32 RGB images of real-world house numbers divided into 10 classes, 73257 for training and 26032 for testing. Mini-ImageNet is a subset of ImageNet with 50K samples for training and 10K for testing divided into 100 classes and down-sampled to 84$\times$84. Unless otherwise stated, all the experiments use standard data augmentation: random cropping with padding of 4 pixels per side and random horizontal flip (except in SVHN, where horizontal flip is omitted).

\vspace{0.2em}
\noindent\textbf{Training details.~}
We train a ResNet-18 architecture \cite{2016_CVPR_ResNet} for 200 epochs with SGD with momentum of 0.9 and a batch size of 128. We use two learning rate schedules: step-wise and linear decay. For both schedules we adopt the budget-aware version proposed by Li et al.~\cite{2020_ICLR_budget} and use an initial learning rate of 0.1. In the step-wise case, the learning rate is divided by 10 at 1/3 (epoch 66) and 2/3 (epoch 133) of the training. The linear schedule decreases the learning rate value at every iteration linearly from the initial value to approximately zero ($10^{-6}$) at the end of the training. We always report the average accuracy and standard deviation of the model across 3 independent runs trained on a GeForce GTX 1080Ti GPU using the Pytorch library. For each budget, we report best results in bold and best results in each section -- data augmentation or learning rate schedule -- in blue (baseline SGD is excluded).

\subsection{Budget-free training for importance sampling}
\begin{wraptable}{r}{0.5\textwidth}
\vskip -0.15in
\caption{Test accuracy (\%), time (min) and speed-up (\%) with respect SGD under a budget-free training (A, T, and S respectively). * denotes that we have used the official code.}\label{tab:t1}
\resizebox{0.5\textwidth}{!}{%
\begin{tabular}{lcccccc}
\toprule 
{} & \multicolumn{3}{c}{CIFAR-10} & \multicolumn{3}{c}{CIFAR-100} \tabularnewline
\midrule 
{Method} &  A & T & S & A  & T  & S \tabularnewline
\midrule 
SGD  & 94.58 {\scriptsize $\pm$ 0.33} & 141  & 0.0 & 74.56 {\scriptsize $\pm$ 0.06} & 141 & 0.0 \tabularnewline
\textit{p-SGD} & 94.41 {\scriptsize $\pm$ 0.19} & 113  & 19.9 & 74.44 {\scriptsize $\pm$ 0.06} & 127 & 9.9  \tabularnewline
\textit{c-SGD}& 94.17 {\scriptsize $\pm$ 0.11} &  100  & 29.1 & 74.40 {\scriptsize $\pm$ 0.06} & 127  & 9.9  \tabularnewline
\textit{SB} (*)& 93.90 {\scriptsize $\pm$ 0.16} &  85   & 39.7 & 73.39 {\scriptsize $\pm$ 0.37} & 119  & 15.6 \tabularnewline 
\textit{OSGD} (*)& 94.34 {\scriptsize $\pm$ 0.07} & 139   & 0.1 & 74.22 {\scriptsize $\pm$ 0.21} & 141  & 0.0 \tabularnewline 
\bottomrule
\end{tabular}
}
\vskip -0.1in
\end{wraptable}

Current importance sampling methods from the state-of-the-art are optimized with no restriction in the number of training iterations. While this allows the methods to better exploit the training process, it makes it difficult to evaluate their computational benefit. Therefore, Table~\ref{tab:t1} presents the performance, wall-clock time, and speed-up relative to a full training of the methods presented in Section~\ref{sec:BT}. All methods train with a step-wise linear learning rate schedule. SGD corresponds to a standard training as described in Subsection~\ref{sub_sec:exp_framework}. \textit{p-SGD} and \textit{c-SGD} correspond to the methods described in Section~\ref{sec:BT} introduced by Chang et al.~\cite{2017_NeurIPS_activeBias} that for the experiments in Table~\ref{tab:t1} train for 200 epochs where the first 70 epochs consist of a warm-up stage with a uniform sampling strategy as done in the original paper. For CIFAR-10 we use a budget of 0.8 for \textit{p-SGD} and 0.7 for \textit{c-SGD}, and for CIFAR-100 a budget of 0.9 for both approaches (budgets retaining most accuracy were selected). Finally, \textit{SB} and \textit{OSGD} follow the setups described in the corresponding papers, \cite{2019_Arxiv_selective} and \cite{2020_AISTATS_osgd}, and run on the official code.

While the simpler approaches to importance sampling, \textit{p-SGD} and \textit{c-SGD}, achieve similar performance to SGD and reduce computation up to 29.08\% (9.93\%) in CIFAR-10 (CIFAR-100), \textit{SB} reduces the training time 39.72\% (15.60\%) in CIFAR-10 (CIFAR-100) with very small drops in accuracy. 
This supports importance sampling observations where particular configurations effectively reduce computational requirements and maintain accuracy.

\subsection{Budgeted training for importance sampling}
%
\begin{wraptable}{r}{0.6\textwidth}
\vskip -0.15in
\caption{\label{tab:t2} Test accuracy with a step-wise and a linear learning rate decay under different budgets. Note that \textit{SB} requires additional computation (forward passes). }
\resizebox{0.6\textwidth}{!}{%
\begin{tabular}{lcccccc}
\toprule 
{} & \multicolumn{3}{c}{CIFAR-10} & \multicolumn{3}{c}{CIFAR-100}\tabularnewline
\midrule 
\textit{SGD - SLR}  & \multicolumn{3}{c}{94.58 $\pm$ 0.33} &  \multicolumn{3}{c}{74.56 $\pm$ 0.06}\tabularnewline
\textit{SGD - LLR}  & \multicolumn{3}{c}{94.80 $\pm$ 0.08} &  \multicolumn{3}{c}{75.44 $\pm$ 0.16}\tabularnewline
\midrule 
{Budget:} & 0.2 & 0.3  & 0.5 & 0.2 & 0.3  & 0.5 \tabularnewline
\midrule 
{} & \multicolumn{6}{c}{Step-wise decay of the learning rate (SLR)} \tabularnewline
\midrule 
\textit{scan-SGD}  &  92.03 {\scriptsize $\pm$ 0.24}  & 93.06 {\scriptsize $\pm$ 0.15} & 93.80 {\scriptsize $\pm$ 0.15} & 70.89 {\scriptsize $\pm$ 0.23} & \ctB{72.31 {\scriptsize $\pm$ 0.22}} & \ctB{73.49 {\scriptsize $\pm$ 0.20}} \tabularnewline
\textit{unif-SGD} & 91.82 {\scriptsize $\pm$ 0.05}  & 92.69 {\scriptsize $\pm$ 0.07} & 93.71 {\scriptsize $\pm$ 0.07}  & 70.36 {\scriptsize $\pm$ 0.30} & 72.03 {\scriptsize $\pm$ 0.47} & 73.36 {\scriptsize $\pm$ 0.20} \tabularnewline
\textit{p-SGD}  & 92.28 {\scriptsize $\pm$ 0.05} & 92.91 {\scriptsize $\pm$ 0.18} & 93.85 {\scriptsize $\pm$ 0.07}  & 70.24 {\scriptsize $\pm$ 0.28} & 72.11 {\scriptsize $\pm$ 0.39} & 72.94 {\scriptsize $\pm$ 0.36} \tabularnewline
\textit{c-SGD} & 91.70 {\scriptsize $\pm$ 0.25} & 92.83 {\scriptsize $\pm$ 0.30} & 93.71 {\scriptsize $\pm$ 0.15} & 69.86 {\scriptsize $\pm$ 0.36} & 71.56 {\scriptsize $\pm$ 0.27} & 73.02 {\scriptsize $\pm$ 0.34} \tabularnewline
\textit{SB} & \ctB{93.37 {\scriptsize $\pm$ 0.11}} &  \ctB{93.86 {\scriptsize $\pm$ 0.27}}  & \ctB{94.21 {\scriptsize $\pm$ 0.13}} & \ctB{70.94 {\scriptsize $\pm$ 0.38}} & 72.25 {\scriptsize $\pm$ 0.68} & 73.39 {\scriptsize $\pm$ 0.37} \tabularnewline 
\textit{OSGD} & 90.61 {\scriptsize $\pm$ 0.31} & 91.78 {\scriptsize $\pm$ 0.30} & 93.45 {\scriptsize $\pm$ 0.10} & 70.09 {\scriptsize $\pm$ 0.25} & 72.18 {\scriptsize $\pm$ 0.35} & 73.39 {\scriptsize $\pm$ 0.22} \tabularnewline 
\midrule 
{} & \multicolumn{6}{c}{Linear decay of the learning rate (LLR)} \tabularnewline
\midrule 
\textit{scan-SGD}  & 92.95 {\scriptsize $\pm$ 0.07} & 93.55 {\scriptsize $\pm$ 0.21} & 94.22 {\scriptsize $\pm$ 0.16} & \textbf{72.04 {\scriptsize $\pm$ 0.42}} & 72.97 {\scriptsize $\pm$ 0.07} & 73.90 {\scriptsize $\pm$ 0.43} \tabularnewline
\textit{unif-SGD} & 92.83 {\scriptsize $\pm$ 0.14}  & 93.48 {\scriptsize $\pm$ 0.05} & 93.98 {\scriptsize $\pm$ 0.11}  & 72.02 {\scriptsize $\pm$ 0.24} & 72.74 {\scriptsize $\pm$ 0.57} & 73.93 {\scriptsize $\pm$ 0.16}  \tabularnewline
\textit{p-SGD}  & 93.23 {\scriptsize $\pm$ 0.14}  & 93.63 {\scriptsize $\pm$ 0.04} & 94.14 {\scriptsize $\pm$ 0.11} & 71.72 {\scriptsize $\pm$ 0.37}  & 72.94 {\scriptsize $\pm$ 0.37} & 74.06 {\scriptsize $\pm$ 0.10} \tabularnewline
\textit{c-SGD} & 92.95 {\scriptsize $\pm$ 0.17} & 93.54 {\scriptsize $\pm$ 0.07} & 94.11 {\scriptsize $\pm$ 0.24} & 71.37 {\scriptsize $\pm$ 0.49} & 72.33 {\scriptsize $\pm$ 0.18} & 73.93 {\scriptsize $\pm$ 0.35} \tabularnewline
\textit{SB} & \textbf{93.78 {\scriptsize $\pm$ 0.11}} & \textbf{94.06 {\scriptsize $\pm$ 0.37}} & \textbf{94.57 {\scriptsize $\pm$ 0.18}} & 71.96 {\scriptsize $\pm$ 0.67} & \textbf{73.11 {\scriptsize $\pm$ 0.42}}  & \textbf{74.35 {\scriptsize $\pm$ 0.34}}  \tabularnewline 
\textit{OSGD} & 91.87 {\scriptsize $\pm$ 0.36} & 93.00 {\scriptsize $\pm$ 0.08} & 93.93 {\scriptsize $\pm$ 0.22} & 71.25 {\scriptsize $\pm$ 0.11} & 72.56 {\scriptsize $\pm$ 0.36} & 73.40 {\scriptsize $\pm$ 0.14} \tabularnewline 
\bottomrule
\end{tabular}
}
\vskip -0.2in
\end{wraptable}
We adapt importance sampling approaches as described in Section~\ref{sec:BT} and configure each method to constrain its computation to the given budget. Table~\ref{tab:t2} shows the analyzed methods performance under the same budget for a step-wise learning rate (SLR) decay and the linear decay (LLR) proposed by Li et al.~\cite{2020_ICLR_budget} for budgeted training (described in Section~\ref{sub_sec:exp_framework}). Surprisingly, this setup shows that most methods achieve very similar performance given a predefined budget, thus not observing faster convergence when using importance sampling. Both \textit{p-SGD} and \textit{c-SGD} provide marginal or no improvements: \textit{p-SGD} marginally improves \textit{unif-SGD} in CIFAR-10, but fails to do so in CIFAR-100. Similar behaviour is observed in \textit{c-SGD}. Conversely, \textit{SB} surpasses the other approaches consistently for SLR and in most cases in the LLR setup. However, \textit{SB} introduces additional forward passes not considered as budget, 
while the other methods do not (see Section~\ref{sec:BT} for an estimation of the budget used by SB). 

We consider \textit{scan-SGD} and \textit{unif-SGD}, as two naive baselines for budgeted training. Despite having similar results (\textit{scan-SGD} seems to be marginally better than \textit{unif-SGD}), we use \textit{unif-SGD} for further experimentation in the following subsections as it adopts a uniform random sampling distribution, which allows contrasting with the importance sampling methods. Additionally, Table~\ref{tab:t2} confirms the effectiveness of a linear learning rate schedule as proposed in \cite{2020_ICLR_budget}: all methods consistently improve with this schedule and, in most cases, \textit{unif-SGD} and LLR perform on par with \textit{SB} and SLR, and surpasses all the other methods when using SLR.

This failure of the sampling strategies to consistently outperform \textit{unif-SGD} could be explained by importance sampling breaking the assumption that samples are i.i.d: SGD assumes that a set of randomly selected samples represents the whole dataset and provides an unbiased estimation of the gradients. Importance sampling explicitly breaks this assumption and biases the gradient estimates. While this might produce gradient estimates that have a bigger impact on the loss, breaking the i.i.d. assumption leads SGD to biased solutions~\cite{2015_ICLRw_online, 2017_NeurIPS_activeBias, 2019_NeurIPS_autoAssist}, which offsets the possible benefits of training with the most relevant samples. As a result, importance sampling does not bring a consistent speed-up in training. Note that approaches that weight the contribution of each sample with the inverse sampling probability to generate an unbiased gradient estimate obtain similar results~\cite{2016_ICLR_varianceRed, 2016_ICML_adaptive, 2017_NeurIPS_activeBias, 2018_NeurIPS_RAIS, 2019_NeurIPS_autoAssist}.

\subsection{Data variability importance during training}\label{sub_sec:EXP_VAR}

\begin{figure}
\centering{}%
\begin{tabular}{cccc}
\hskip -0.1in
\includegraphics[width=0.24\textwidth,height=0.104\textheight]{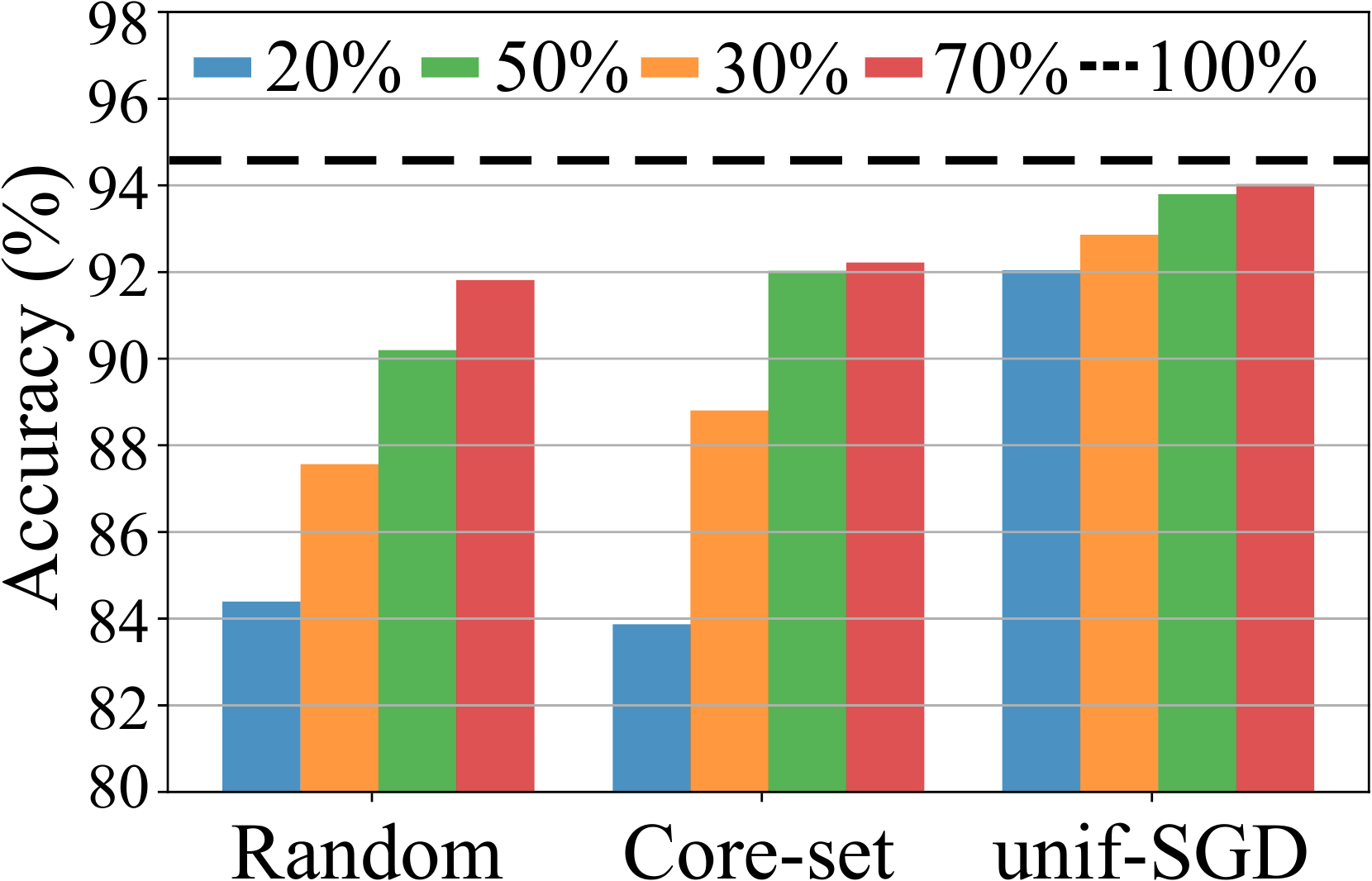} &
\hskip -0.1in
\includegraphics[width=0.24\textwidth,height=0.104\textheight]{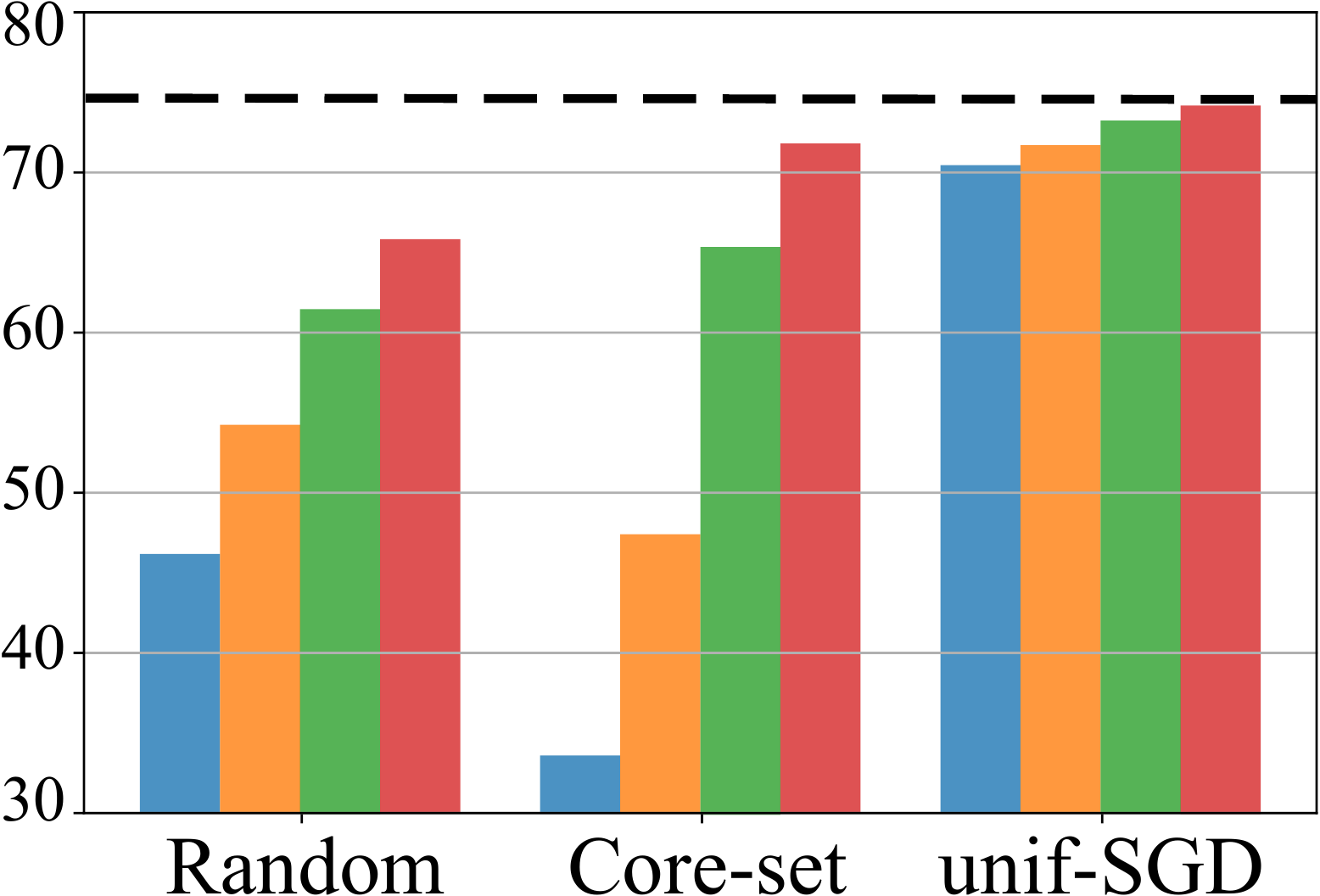} &
\hskip -0.1in
\includegraphics[width=0.24\textwidth,height=0.110\textheight]{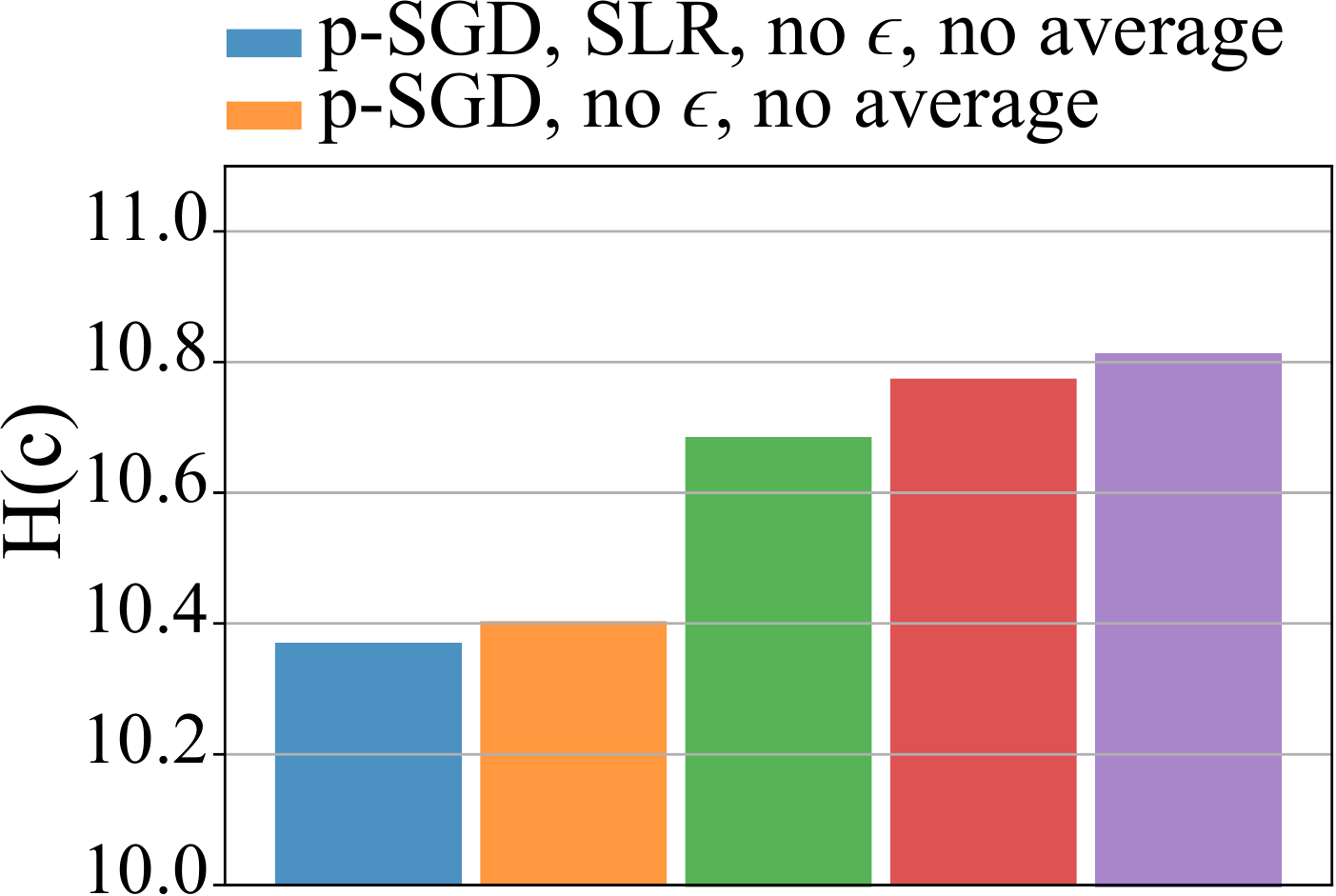} &
\hskip -0.1in
\includegraphics[width=0.24\textwidth,height=0.110\textheight]{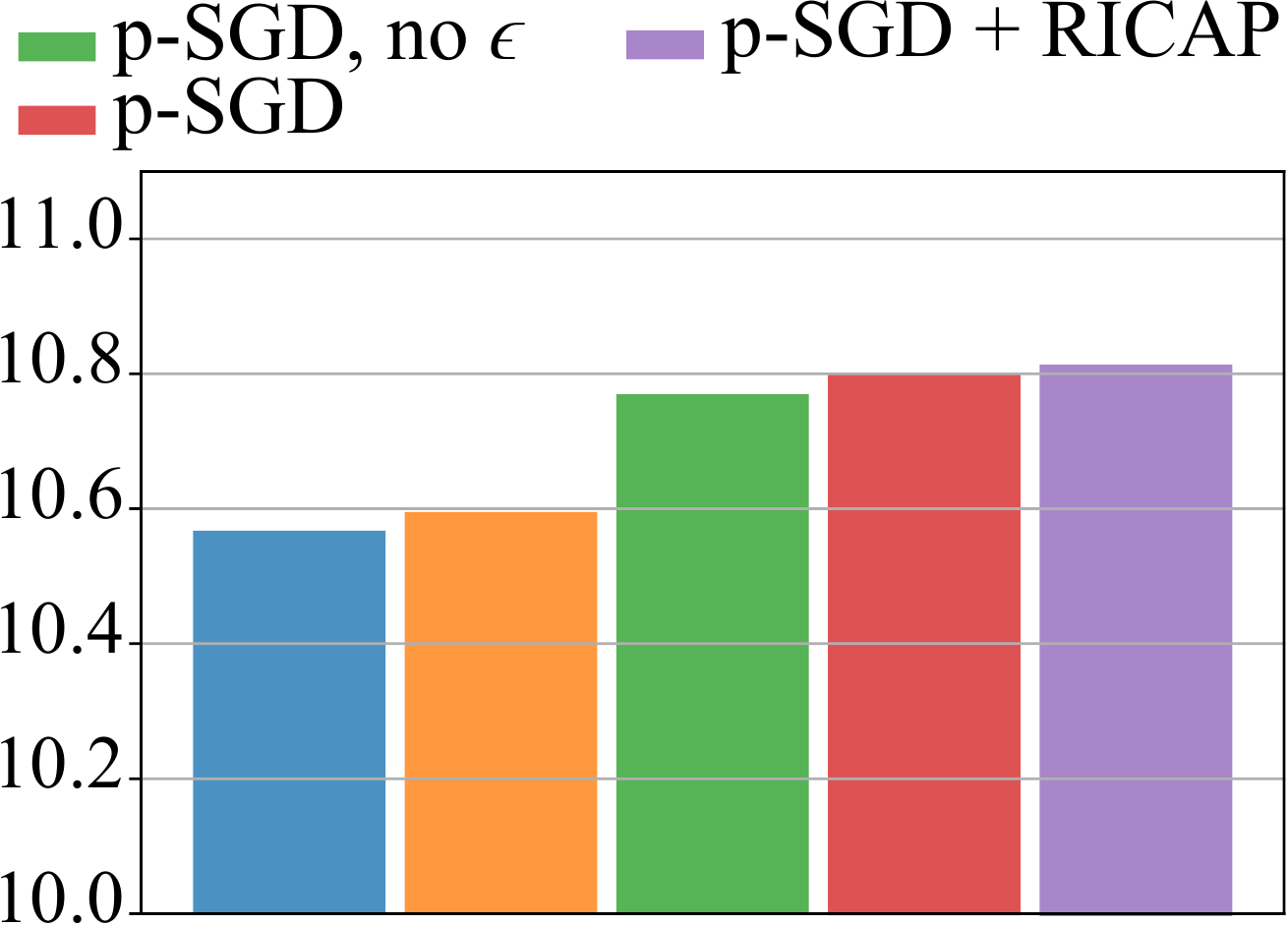}
\\  (a)&(b)&(c)&(d)
\end{tabular}
\caption{\label{fig1_coreset} Importance of data variability in CIFAR-10, (a) and (c), and CIFAR-100, (b) and (d). (a) and (b) compare different training set selection strategies: randomly selecting samples at every epoch (\textit{unif-SGD}) outperforms fixed core-set or random subsets. (c) and (d) compares the data variability of different training strategies: the entropy of sample counts during training (0.3 budget) demonstrates that importance sampling, linear learning rate, and data augmentation contribute to higher data variability (entropy).}
\vskip -0.2in
\end{figure}

Core-set selection approaches~\cite{2018_ICLR_forget, 2020_ICLR_proxySelection} aim to find the most representative samples in the dataset to make training more efficient, while keeping accuracy as high as possible. Figure~\ref{fig1_coreset}, (a) and (b), presents how core-set selection and a randomly chosen subset (Random) both under-perform uniform random sampling of a subset each epoch (\textit{unif-SGD}), which approaches standard training performance (black dashed line). This shows that randomly selecting a different subset every epoch (\textit{unif-SGD}), which is equally computationally efficient, achieves substantially better accuracy. This result supports the widely adopted assumption that data variability is key and suggests that it might be more important than sample quality.

We also find data variability to play an important role within importance sampling. Figure~\ref{fig1_coreset} (c) and (d) shows data variability measured using the entropy $H(c)$ of the number of times that a sample is seen by the network during training, with $c$ being the $N$-D distribution of sample counts. 
These results show how increases in variability (higher entropy) follow accuracy improvements in \textit{p-SGD} when introducing the LLR, the smoothing constant to the $P^{t}$ sampling distribution, the average of the predictions across epochs, and data augmentation.

\subsection{Data augmentation for importance sampling}\label{sub_sec:EXP_DA}
%
%
\begin{wraptable}{r}{0.6\textwidth}
\vskip -0.15in
\caption{\label{tab:t3} Data augmentation for budgeted importance sampling in CIFAR-10/100. N and M are the number and strength of RandAugment augmentations, and $\alpha$ controls the interpolation in mixup and RICAP. Note that SGD corresponds to the full training. }
\resizebox{0.6\textwidth}{!}{%
\begin{tabular}{lcccccc}
\toprule 
{}  & \multicolumn{3}{c}{CIFAR-10} & \multicolumn{3}{c}{CIFAR-100} \tabularnewline
\midrule 
{Budget:} & 0.2 & 0.3  & 0.5 & 0.2 & 0.3  & 0.5 \tabularnewline
\midrule 
{} & \multicolumn{6}{c}{Standard data augmentation}\tabularnewline
\midrule 
SGD ($B = 1$) & \multicolumn{3}{c}{94.80 {\scriptsize $\pm$ 0.08}} &  \multicolumn{3}{c}{75.44 {\scriptsize $\pm$ 0.16}}\tabularnewline
\textit{unif-SGD} & 92.83 {\scriptsize $\pm$ 0.14} & 93.48 {\scriptsize $\pm$ 0.05} & 93.98 {\scriptsize $\pm$ 0.11}  & \ctB{72.02 {\scriptsize $\pm$ 0.24}} & 72.74 {\scriptsize $\pm$ 0.57} & 73.93 {\scriptsize $\pm$ 0.16}   \tabularnewline
\textit{p-SGD}  &   93.23 {\scriptsize $\pm$ 0.14} & 93.63 {\scriptsize $\pm$ 0.04} & 94.14 {\scriptsize $\pm$ 0.11} & 71.72 {\scriptsize $\pm$ 0.37}  & 72.94 {\scriptsize $\pm$ 0.37} & 74.06 {\scriptsize $\pm$ 0.10}  \tabularnewline
\textit{SB} & \ctB{93.78 {\scriptsize $\pm$ 0.11}} & \ctB{94.06 {\scriptsize $\pm$ 0.37}} & \ctB{94.57 {\scriptsize $\pm$ 0.18}} & 71.96 {\scriptsize $\pm$ 0.67} & \ctB{73.11 {\scriptsize $\pm$ 0.42}}  & \ctB{74.35 {\scriptsize $\pm$ 0.34}}  \tabularnewline 
\midrule 
{} & \multicolumn{6}{c}{RandAugment data augmentation (N = 2, M = 4)}\tabularnewline
\midrule 
SGD ($B = 1$) & \multicolumn{3}{c}{95.56 {\scriptsize$\pm$ 0.12}} &  \multicolumn{3}{c}{75.52 {\scriptsize$\pm$ 0.17}}\tabularnewline
\textit{unif-SGD} & 92.76 {\scriptsize$\pm$ 0.16} & 93.78 {\scriptsize$\pm$ 0.11} &  94.64 {\scriptsize$\pm$ 0.08} & 71.44 {\scriptsize$\pm$ 0.37} & \ctB{73.23 {\scriptsize$\pm$ 0.29}} & 74.78 {\scriptsize$\pm$ 0.45} \tabularnewline
\textit{p-SGD} & 92.95 {\scriptsize$\pm$ 0.31} & 93.99 {\scriptsize$\pm$ 0.28} & 94.91 {\scriptsize$\pm$ 0.18} & \ctB{71.63 {\scriptsize$\pm$ 0.27}} & 72.91 {\scriptsize$\pm$ 0.13} & 74.30 {\scriptsize$\pm$ 0.04} \tabularnewline
\textit{SB} & \ctB{93.27 {\scriptsize$\pm$ 0.38}} & \ctB{94.64 {\scriptsize$\pm$ 0.07}} &  \ctB{95.27 {\scriptsize$\pm$ 0.26}} & 66.84 {\scriptsize$\pm$ 1.15} & 73.79 {\scriptsize$\pm$ 0.40} & \ctB{74.87 {\scriptsize$\pm$ 0.18}} \tabularnewline
\midrule 
{} & \multicolumn{6}{c}{mixup data augmentation ($\alpha$ = 0.3)}\tabularnewline
\midrule 
SGD ($B = 1$) & \multicolumn{3}{c}{95.82 {\scriptsize $\pm$ 0.17}} &  \multicolumn{3}{c}{77.62 {\scriptsize $\pm$ 0.40}}\tabularnewline
\textit{unif-SGD} & 93.64 {\scriptsize $\pm$ 0.27} & \ctB{94.49 {\scriptsize $\pm$ 0.04}} &  95.18 {\scriptsize $\pm$ 0.05} & 73.28 {\scriptsize $\pm$ 0.51} & \ctB{75.13 {\scriptsize $\pm$ 0.52}} & 75.80 {\scriptsize $\pm$ 0.34} \tabularnewline
\textit{p-SGD} & \ctB{93.78 {\scriptsize $\pm$ 0.04}} & 94.41 {\scriptsize $\pm$ 0.16} &  \ctB{95.26 {\scriptsize $\pm$ 0.06}} & 73.35 {\scriptsize $\pm$ 0.29}& 75.05 {\scriptsize $\pm$ 0.15} & \ctB{75.87 {\scriptsize $\pm$ 0.15}} \tabularnewline
\textit{SB} & 93.62 {\scriptsize $\pm$ 0.36} & 93.92 {\scriptsize $\pm$ 0.08} &  94.51 {\scriptsize $\pm$ 0.17} & \ctB{73.38 {\scriptsize $\pm$ 0.13}} & 74.88 {\scriptsize $\pm$ 0.31} & 75.57 {\scriptsize $\pm$ 0.23} \tabularnewline %
\midrule 
{} & \multicolumn{6}{c}{RICAP data augmentation ($\alpha$ = 0.3)}\tabularnewline
\midrule 
SGD ($B = 1$) & \multicolumn{3}{c}{96.17 {\scriptsize $\pm$ 0.09}} &  \multicolumn{3}{c}{78.91 {\scriptsize $\pm$ 0.07}}\tabularnewline
\textit{unif-SGD} & 93.85 {\scriptsize $\pm$ 0.10}  & \textbf{94.93 {\scriptsize $\pm$ 0.29}}  &  95.47 {\scriptsize $\pm$ 0.18} & \textbf{74.87 {\scriptsize $\pm$ 0.28}} & 76.27 {\scriptsize $\pm$ 0.32}  & \textbf{77.83 {\scriptsize $\pm$ 0.15}}  \tabularnewline
\textit{p-SGD} & \textbf{94.02 {\scriptsize $\pm$ 0.18}}  & 94.79 {\scriptsize $\pm$ 0.18}  &  \textbf{95.63 {\scriptsize $\pm$ 0.15}} & 74.59 {\scriptsize $\pm$ 0.15} & \textbf{76.50 {\scriptsize $\pm$ 0.22}}  & 77.58 {\scriptsize $\pm$ 0.49}  \tabularnewline
\textit{SB} & 89.93 {\scriptsize $\pm$ 0.84} & 93.64 {\scriptsize $\pm$ 0.42} & 94.76 {\scriptsize $\pm$ 0.02} & 56.66 {\scriptsize $\pm$ 0.65} & 72.24 {\scriptsize $\pm$ 0.58} & 76.26 {\scriptsize $\pm$ 0.22} \tabularnewline %
\bottomrule
\end{tabular}
}
\vskip -0.15in
\end{wraptable}
Importance sampling approaches usually do not explore the interaction of sampling strategies with data augmentation techniques \cite{2015_ICLRw_online, 2018_ICML_notAllSamples, 2019_Arxiv_selective}. To better understand this interaction, we explore interpolation-based augmentations via RICAP \cite{2018_ACML_RICAP} and mixup \cite{2018_ICLR_mixup}; and non-interpolation augmentations using RandAugment \cite{2020_CVPRw_randAugm}. We implemented these data augmentation policies as reported in the original papers (see Table~\ref{tab:t3} for the hyperparameters used in our experiments). Note that for mixup and RICAP we combine 2 and 4 images respectively within each mini-batch, which results in the same number of samples being shown to the network ($T \times B$).

%
\begin{wraptable}{r}{0.6\textwidth}
\caption{\label{tab:t4} Data augmentation for budgeted importance sampling in SVHN and mini-ImageNet. N and M are the number and strength of RandAugment augmentations, and $\alpha$ controls the interpolation in mixup and RICAP. }
\resizebox{0.6\textwidth}{!}{%
\begin{tabular}{lcccccc}
\toprule 
{}  & \multicolumn{3}{c}{SVHN} & \multicolumn{3}{c}{mini-ImageNet} \tabularnewline
\midrule 
{Budget:} & 0.2 & 0.3 & 0.5 & 0.2 & 0.3 & 0.5  \tabularnewline
\midrule 
{} & \multicolumn{6}{c}{Standard data augmentation}\tabularnewline
\midrule 
SGD ($B = 1$) & \multicolumn{3}{c}{97.02 {\scriptsize $\pm$ 0.05}} &  \multicolumn{3}{c}{75.19 {\scriptsize $\pm$ 0.16}}\tabularnewline
\textit{unif-SGD} & 96.56 {\scriptsize $\pm$ 0.12} & 96.78 {\scriptsize $\pm$ 0.13} & 96.95 {\scriptsize $\pm$ 0.07} & 70.87 {\scriptsize $\pm$ 0.56} & 72.19 {\scriptsize $\pm$ 0.43} & \ctB{73.88 {\scriptsize $\pm$ 0.42}} \tabularnewline
\textit{p-SGD} & 96.52 {\scriptsize $\pm$ 0.03} & 96.75 {\scriptsize $\pm$ 0.03} & 96.84 {\scriptsize $\pm$ 0.06} & \ctB{71.05 {\scriptsize $\pm$ 0.29}} & \ctB{72.39 {\scriptsize $\pm$ 0.45}} & 73.66 {\scriptsize $\pm$ 0.39} \tabularnewline
\textit{SB} & \ctB{96.93 {\scriptsize $\pm$ 0.07}} & \ctB{96.85 {\scriptsize $\pm$ 0.01}} & \ctB{96.97 {\scriptsize $\pm$ 0.06}} & 69.68 {\scriptsize $\pm$ 0.09} & 71.46 {\scriptsize $\pm$ 0.15} & 73.51 {\scriptsize $\pm$ 0.30} \tabularnewline
\midrule 
{} & \multicolumn{6}{c}{RandAugment data augmentation (N = 2, M = 4)}\tabularnewline
\midrule 
SGD ($B = 1$) & \multicolumn{3}{c}{97.59 {\scriptsize $\pm$ 0.14}} &  \multicolumn{3}{c}{74.15 {\scriptsize $\pm$ 0.22}}\tabularnewline
\textit{unif-SGD} & 97.38 {\scriptsize $\pm$ 0.05} & \ctB{97.50 {\scriptsize $\pm$ 0.07}} & \ctB{97.60 {\scriptsize $\pm$ 0.05}} & 71.29 {\scriptsize $\pm$ 0.25} & \ctB{73.04 {\scriptsize $\pm$ 0.34}} & 73.21 {\scriptsize $\pm$ 0.52} \tabularnewline
\textit{p-SGD} & 97.25 {\scriptsize $\pm$ 0.03} & 97.44 {\scriptsize $\pm$ 0.02} & 97.52 {\scriptsize $\pm$ 0.03} & \ctB{71.43 {\scriptsize $\pm$ 0.25}} & 72.36 {\scriptsize $\pm$ 0.15} & 73.21 {\scriptsize $\pm$ 0.38} \tabularnewline
\textit{SB} & \ctB{97.42 {\scriptsize $\pm$ 0.09}} & 97.43 {\scriptsize $\pm$ 0.19} & 97.56 {\scriptsize $\pm$ 0.05} & 67.17 {\scriptsize $\pm$ 2.51} & 71.69 {\scriptsize $\pm$ 0.31} & \ctB{73.28 {\scriptsize $\pm$ 0.03}} \tabularnewline
\midrule 
{} & \multicolumn{6}{c}{mixup data augmentation ($\alpha$ = 0.3)}\tabularnewline
\midrule 
SGD ($B = 1$) & \multicolumn{3}{c}{97.24 {\scriptsize $\pm$ 0.03}} &  \multicolumn{3}{c}{76.28 {\scriptsize $\pm$ 0.28}}\tabularnewline
\textit{unif-SGD} & \ctB{96.99 {\scriptsize $\pm$ 0.09}} & 97.04 {\scriptsize $\pm$ 0.08} & 97.24 {\scriptsize $\pm$ 0.07} & \ctB{72.50 {\scriptsize $\pm$ 0.51}} & \ctB{73.76 {\scriptsize $\pm$ 0.26}} & \ctB{75.05 {\scriptsize $\pm$ 0.29}} \tabularnewline
\textit{p-SGD} & 96.92 {\scriptsize $\pm$ 0.08} & \ctB{97.34 {\scriptsize $\pm$ 0.49}} & \ctB{97.37 {\scriptsize $\pm$ 0.49}} & 72.21 {\scriptsize $\pm$ 0.81} & 73.63 {\scriptsize $\pm$ 0.13} & 74.54 {\scriptsize $\pm$ 0.53} \tabularnewline
\textit{SB} & 96.80 {\scriptsize $\pm$ 0.09} & 96.92 {\scriptsize $\pm$ 0.09} & 96.96 {\scriptsize $\pm$ 0.09} & 70.12 {\scriptsize $\pm$ 0.51} & 72.01 {\scriptsize $\pm$ 0.72} & 73.76 {\scriptsize $\pm$ 0.36} \tabularnewline
\midrule 
{} & \multicolumn{6}{c}{RICAP data augmentation ($\alpha$ = 0.3)}\tabularnewline
\midrule 
SGD ($B = 1$) & \multicolumn{3}{c}{97.61 {\scriptsize $\pm$ 0.06}} &  \multicolumn{3}{c}{78.75 {\scriptsize $\pm$ 0.40}}\tabularnewline
\textit{unif-SGD} & 97.47 {\scriptsize $\pm$ 0.04} & \textbf{97.62 {\scriptsize $\pm$ 0.16}} & 97.55 {\scriptsize $\pm$ 0.04} & 73.56 {\scriptsize $\pm$ 0.24} & 75.15 {\scriptsize $\pm$ 0.45} & 77.20 {\scriptsize $\pm$ 0.33} \tabularnewline
\textit{p-SGD} & \textbf{97.48 {\scriptsize $\pm$ 0.08}} & 97.45 {\scriptsize $\pm$ 0.06} & \textbf{97.57 {\scriptsize $\pm$ 0.05}} & \textbf{73.67 {\scriptsize $\pm$ 0.60}} & \textbf{75.46 {\scriptsize $\pm$ 0.27}} & \textbf{77.25 {\scriptsize $\pm$ 0.47}} \tabularnewline
\textit{SB} & 97.34 {\scriptsize $\pm$ 0.03} & 97.40 {\scriptsize $\pm$ 0.06} & 97.45 {\scriptsize $\pm$ 0.01} & 53.26 {\scriptsize $\pm$ 0.71} & 71.75 {\scriptsize $\pm$ 0.67} & 75.65 {\scriptsize $\pm$ 0.40} \tabularnewline
\bottomrule
\end{tabular}
}
\vskip -0.1in
\end{wraptable}
Table~\ref{tab:t3} and~\ref{tab:t4} show that data augmentation is beneficial in a budgeted training scenario: in most cases all strategies increase performance compared to standard data augmentation. The main exception is for the lowest budget for \textit{SB} where in some cases data augmentation hurts performance.
In particular, with RICAP and mixup, the improvements from importance sampling approaches are marginal and the naive \textit{unif-SGD} provides results close to full training with standard augmentation. In some cases \textit{unif-SGD} surpasses full-training with standard augmentations, e.g.~RICAP with 0.3 and 0.5 budget and both mixup and RICAP with 0.3 budget in CIFAR-10/100. This is even more evident in SVHN where all the budgets in Table~\ref{tab:t4} for \textit{unif-SGD} with RICAP surpass full training (SGD) with standard augmentation.

\begin{wraptable}{r}{0.5\textwidth}
\vskip -0.15in
\caption{\label{tab:t6_wallClock} Wall-clock time (minutes) in CIFAR-100 for a training of 0.3 of budget.}
\resizebox{0.5\textwidth}{!}{%
{
\begin{tabular}{lccc}
\toprule 
{Approaches:} & \textit{unif-SGD} & \textit{p-SGD} & \textit{SB} \tabularnewline
\midrule 
Standard data augmentation & 47 & 48 & 91 \tabularnewline
RandAugment & 48 & 48 & 93 \tabularnewline
mixup & 48 & 48 & 93 \tabularnewline
RICAP & 49 & 49 & 95 \tabularnewline
\bottomrule
\end{tabular}
}
}
\vskip -0.1in
\end{wraptable}
Given that the cost of the data augmentation policies used is negligible (see Table~\ref{tab:t6_wallClock} for the wall-clock times when $B=0.3$), our results show that adequate data augmentation alone can reduce training time at no accuracy cost and in some cases with a considerable increase in accuracy. For example, a 70\% reduction in training time (0.3 budget) corresponds to an increase in accuracy from 75.44\% to 76.27\% in CIFAR-100 and from 94.80\% to 94.93\% in CIFAR-10. Also, a 50\% reduction (0.5 budget) corresponds to an increase in accuracy from 75.44\% to 77.83\% in CIFAR-100 and from 94.80\% to 95.47\% in CIFAR-10.

%
\begin{wraptable}{r}{0.6\textwidth}
\vskip -0.15in
\caption{\label{tab:t5_extreme} Test accuracy for CIFAR-10/100 and mini-ImageNet under extreme budgets. }
\resizebox{0.6\textwidth}{!}{%
\begin{tabular}{lcccccc}
\toprule 
{}  & \multicolumn{2}{c}{CIFAR-10} &  \multicolumn{2}{c}{CIFAR-100} &  \multicolumn{2}{c}{mini-ImageNet} \tabularnewline
\midrule 
{Budget:} & 0.05 & 0.1 & 0.05 & 0.1 & 0.05 & 0.1 \tabularnewline
\midrule 
{} & \multicolumn{6}{c}{Standard data augmentation}\tabularnewline
\midrule 
\textit{unif-SGD}  & 87.90 {\scriptsize $\pm$ 0.40}  & 91.46 {\scriptsize $\pm$ 0.08} & \textbf{62.66 {\scriptsize $\pm$ 0.65}}  & \ctB{69.34 {\scriptsize $\pm$ 0.68}}  & 56.38 {\scriptsize $\pm$ 0.11} & 67.61 {\scriptsize $\pm$ 0.52} \tabularnewline
\textit{p-SGD}  & \textbf{88.86 {\scriptsize $\pm$ 0.17}}  & 91.66 {\scriptsize $\pm$ 0.11} & 62.20 {\scriptsize $\pm$ 0.56}  & 69.32 {\scriptsize $\pm$ 0.17}  & \textbf{56.95 {\scriptsize $\pm$ 0.43}} & \ctB{67.67 {\scriptsize $\pm$ 0.41}} \tabularnewline
\textit{SB} & 79.45 {\scriptsize $\pm$ 4.31} & \textbf{92.66 {\scriptsize $\pm$ 0.14}} & 50.53 {\scriptsize $\pm$ 2.27}  & 68.29 {\scriptsize $\pm$ 0.68}  & 11.19 {\scriptsize $\pm$ 3.46} & 61.25 {\scriptsize $\pm$ 1.76} \tabularnewline
\midrule 
{} & \multicolumn{6}{c}{RandAugment data augmentation (N = 2, M = 4)}\tabularnewline
\midrule 
\textit{unif-SGD}  & 83.24 {\scriptsize $\pm$ 0.06} & 88.95 {\scriptsize $\pm$ 0.22} & 47.64 {\scriptsize $\pm$ 3.34} & 64.48 {\scriptsize $\pm$ 0.10} & \ctB{42.35 {\scriptsize $\pm$ 1.54}} & 64.98 {\scriptsize $\pm$ 0.47} \tabularnewline
\textit{p-SGD} & \ctB{83.94 {\scriptsize $\pm$ 0.26}} & \ctB{89.77 {\scriptsize $\pm$ 0.38}} & \ctB{48.78 {\scriptsize $\pm$ 1.48}} & \ctB{65.05 {\scriptsize $\pm$ 0.37}} & 41.72 {\scriptsize $\pm$ 0.77} & \ctB{65.88 {\scriptsize $\pm$ 0.15}} \tabularnewline
\textit{SB} & 32.21 {\scriptsize $\pm$ 4.14}  & 33.86 {\scriptsize $\pm$ 5.02} & 5.05 {\scriptsize $\pm$ 0.64}  & 5.05 {\scriptsize $\pm$ 0.64}  & 5.61 {\scriptsize $\pm$ 0.66} & 5.94 {\scriptsize $\pm$ 0.13} \tabularnewline
\midrule 
{} & \multicolumn{6}{c}{mixup data augmentation ($\alpha$ = 0.3)}\tabularnewline
\midrule 
\textit{unif-SGD}  & 87.33 {\scriptsize $\pm$ 0.42}  & 91.74 {\scriptsize $\pm$ 0.04} & \ctB{59.90 {\scriptsize $\pm$ 0.71}}  & \textbf{70.43 {\scriptsize $\pm$ 0.45}}  & 53.13 {\scriptsize $\pm$ 0.83} & \textbf{68.54 {\scriptsize $\pm$ 0.98}} \tabularnewline
\textit{p-SGD}  & \ctB{87.56 {\scriptsize $\pm$ 0.67}}  & 91.59 {\scriptsize $\pm$ 0.17} & 59.68 {\scriptsize $\pm$ 0.71}  & 70.31 {\scriptsize $\pm$ 0.10}  & \ctB{54.20 {\scriptsize $\pm$ 0.95}} & 68.39 {\scriptsize $\pm$ 0.46} \tabularnewline
\textit{SB}  & 77.72 {\scriptsize $\pm$ 5.31}  & \ctB{92.56 {\scriptsize $\pm$ 0.15}} & 43.27 {\scriptsize $\pm$ 7.37}  & 69.64 {\scriptsize $\pm$ 0.24}  & 12.10 {\scriptsize $\pm$ 0.27} & 61.01 {\scriptsize $\pm$ 0.64} \tabularnewline
\midrule 
{} & \multicolumn{6}{c}{RICAP data augmentation ($\alpha$ = 0.3)}\tabularnewline
\midrule 
\textit{unif-SGD} & \ctB{85.61 {\scriptsize $\pm$ 0.24}} & \ctB{91.32 {\scriptsize $\pm$ 0.28}} & 55.85 {\scriptsize $\pm$ 0.51} & 69.43 {\scriptsize $\pm$ 0.33} & 48.95 {\scriptsize $\pm$ 0.65} & 67.26 {\scriptsize $\pm$ 0.63} \tabularnewline
\textit{p-SGD} & 85.57 {\scriptsize $\pm$ 0.70} & 90.94 {\scriptsize $\pm$ 0.16} & \ctB{56.09 {\scriptsize $\pm$ 0.71}} & \ctB{70.05 {\scriptsize $\pm$ 0.07}} & \ctB{49.35 {\scriptsize $\pm$ 0.60}} & \ctB{67.27 {\scriptsize $\pm$ 0.85}} \tabularnewline
\textit{SB} & 44.93 {\scriptsize $\pm$ 2.67} & 54.76 {\scriptsize $\pm$ 4.31} & 10.75 {\scriptsize $\pm$ 0.72} & 13.33 {\scriptsize $\pm$ 0.39} & 8.71 {\scriptsize $\pm$ 0.45} & 10.84 {\scriptsize $\pm$ 0.86} \tabularnewline
\bottomrule
\end{tabular}
}
\vskip -0.15in
\end{wraptable}
We also experimented with extremely low budgets (see Table~\ref{tab:t5_extreme}) and found that importance sampling approaches (\textit{p-SGD} and \textit{SB}) still bring little improvement over uniform random sampling (\textit{unif-SGD}). Here additional data augmentation does not bring a significant improvement in accuracy and in the most challenging cases, hinders convergence. For example, when introducing RICAP with $B=0.05$, the accuracy drops approximately 2 points in CIFAR-10, 5 points in CIFAR-100, and 7 points in mini-ImageNet with respect to 87.90\%, 62.66\%, and 56.38\% for \textit{unif-SGD} with standard data augmentation.

\section{Conclusion}
This paper studied DNN training for image classification when the number of iterations is fixed (i.e.~budgeted training) and explores the interaction of importance sampling techniques and data augmentation in this setup. We empirically showed that, in budgeted training, DNNs prefer variability over selection of important samples: adequate data augmentation surpasses state-of-the-art importance sampling methods and allows for up to a 70\% reduction of the training time (budget) with no loss (and sometimes an increase) in accuracy. In future work, we plan to explore the limitations found in extreme budgets and extend the study to large-scale datasets where training DNNs becomes a longer process. 
Additionally, we find particularly interesting as future work to study  the generalization of the conclusions presented in this paper to different tasks, types of data, and model architectures.
Finally, we encourage the use of data augmentation techniques rather than importance sampling approaches in scenarios where the iterations budget is restricted, and motivate research on these scenarios to better exploit computational resources.

\section*{Acknowledgment}
This publication has emanated from research conducted with the financial support of Science Foundation Ireland (SFI) under grant number SFI/15/SIRG/3283 and SFI/12/RC/2289\_P2.

\bibliography{budgeted_training_bib}

\begin{thebibliography}{64}
\providecommand{\natexlab}[1]{#1}
\providecommand{\url}[1]{\texttt{#1}}
\expandafter\ifx\csname urlstyle\endcsname\relax
  \providecommand{\doi}[1]{doi: #1}\else
  \providecommand{\doi}{doi: \begingroup \urlstyle{rm}\Url}\fi

\bibitem[Alain et~al.(2016)Alain, Lamb, Sankar, Courville, and
  Bengio]{2016_ICLR_varianceRed}
Guillaume Alain, Alex Lamb, Chinnadhurai Sankar, Aaron Courville, and Yoshua
  Bengio.
\newblock {Variance reduction in sgd by distributed importance sampling}.
\newblock In \emph{{International Conference on Learning Representations
  (ICLR)}}, 2016.

\bibitem[Albert et~al.(2021)Albert, Ortego, Arazo, O'Connor, and
  McGuinness]{2021_IJCNN_relab}
Paul Albert, Diego Ortego, Eric Arazo, Noel~E. O'Connor, and Kevin McGuinness.
\newblock Relab: Reliable label bootstrapping for semi-supervised learning.
\newblock In \emph{International Joint Conference on Neural Networks (IJCNN)},
  2021.

\bibitem[Amiri et~al.(2017)Amiri, Miller, and Savova]{2017_CEMNLP_repeat}
Hadi Amiri, Timothy Miller, and Guergana Savova.
\newblock Repeat before forgetting: Spaced repetition for efficient and
  effective training of neural networks.
\newblock In \emph{Conference on Empirical Methods in Natural Language
  Processing}, 2017.

\bibitem[Arazo et~al.(2019)Arazo, Ortego, Albert, O'Connor, and
  McGuinness]{2019_ICML_BynamicBootstrapping}
Eric Arazo, Diego Ortego, Paul. Albert, Noel O'Connor, and Kevin McGuinness.
\newblock {Unsupervised Label Noise Modeling and Loss Correction}.
\newblock In \emph{{International Conference on Machine Learning (ICML)}},
  2019.

\bibitem[Arazo et~al.(2020)Arazo, Ortego, Albert, O'Connor, and
  McGuinness]{2020_IJCNN_pseudolab}
Eric Arazo, Diego Ortego, Paul Albert, Noel~E O'Connor, and Kevin McGuinness.
\newblock Pseudo-labeling and confirmation bias in deep semi-supervised
  learning.
\newblock In \emph{International Joint Conference on Neural Networks (IJCNN)},
  2020.

\bibitem[Ash et~al.(2020)Ash, Zhang, Krishnamurthy, Langford, and
  Agarwal]{2020_ICLR_deepactivelearning}
Jordan~T Ash, Chicheng Zhang, Akshay Krishnamurthy, John Langford, and Alekh
  Agarwal.
\newblock Deep batch active learning by diverse, uncertain gradient lower
  bounds.
\newblock In \emph{International Conference on Learning Representations
  (ICLR)}, 2020.

\bibitem[Bengio et~al.(2009)Bengio, Louradour, Collobert, and
  Weston]{2009_ICML_CL}
Yoshua Bengio, J{\'e}r{\^o}me Louradour, Ronan Collobert, and Jason Weston.
\newblock Curriculum learning.
\newblock In \emph{International Conference on Machine Learning (ICML)}, 2009.

\bibitem[Berthelot et~al.(2019)Berthelot, Carlini, Goodfellow, Papernot,
  Oliver, and Raffel]{2019_NIPS_MixMatch}
David Berthelot, Nicholas Carlini, Ian Goodfellow, Nicolas Papernot, Avital
  Oliver, and Colin~A Raffel.
\newblock {MixMatch: {A} Holistic Approach to Semi-Supervised Learning}.
\newblock In \emph{{Advances in Neural Information Processing Systems
  (NeurIPS)}}, 2019.

\bibitem[Bottou et~al.(2018)Bottou, Curtis, and Nocedal]{2018_SIAM_optim}
L{\'e}on Bottou, Frank~E Curtis, and Jorge Nocedal.
\newblock Optimization methods for large-scale machine learning.
\newblock \emph{Siam Review}, 2018.

\bibitem[Cai et~al.(2019)Cai, Zhu, and Han]{2019_ICLR_proxylessnas}
Han Cai, Ligeng Zhu, and Song Han.
\newblock {Proxylessnas: Direct neural architecture search on target task and
  hardware}.
\newblock In \emph{{International Conference on Learning Representations
  (ICLR)}}, 2019.

\bibitem[Chang et~al.(2017)Chang, Learned-Miller, and
  McCallum]{2017_NeurIPS_activeBias}
Haw-Shiuan Chang, Erik Learned-Miller, and Andrew McCallum.
\newblock Active bias: Training more accurate neural networks by emphasizing
  high variance samples.
\newblock In \emph{Advances in Neural Information Processing Systems
  (NeurIPS)}, 2017.

\bibitem[Chen et~al.(2019)Chen, Liu, Kira, Wang, and
  Huang]{2019_ICLR_closer_look_to_activeLearning}
Wei-Yu Chen, Yen-Cheng Liu, Zsolt Kira, Yu-Chiang~Frank Wang, and Jia-Bin
  Huang.
\newblock A closer look at few-shot classification.
\newblock In \emph{{International Conference on Learning Representations
  (ICLR)}}, 2019.

\bibitem[Cheng et~al.(2019)Cheng, Lian, Deng, Gao, Tan, and
  Geng]{2019_CVPR_L2G}
Hao Cheng, Dongze Lian, Bowen Deng, Shenghua Gao, Tao Tan, and Yanlin Geng.
\newblock Local to global learning: Gradually adding classes for training deep
  neural networks.
\newblock In \emph{IEEE Conference on Computer Vision and Pattern Recognition
  (CVPR)}, 2019.

\bibitem[Coleman et~al.(2020)Coleman, Yeh, Mussmann, Mirzasoleiman, Bailis,
  Liang, Leskovec, and Zaharia]{2020_ICLR_proxySelection}
Cody Coleman, Christopher Yeh, Stephen Mussmann, Baharan Mirzasoleiman, Peter
  Bailis, Percy Liang, Jure Leskovec, and Matei Zaharia.
\newblock {Selection via proxy: Efficient data selection for deep learning}.
\newblock In \emph{{International Conference on Learning Representations
  (ICLR)}}, 2020.

\bibitem[Cubuk et~al.(2020)Cubuk, Zoph, Shlens, and Le]{2020_CVPRw_randAugm}
Ekin~D Cubuk, Barret Zoph, Jonathon Shlens, and Quoc~V Le.
\newblock Randaugment: Practical automated data augmentation with a reduced
  search space.
\newblock In \emph{IEEE Conference on Computer Vision and Pattern Recognition
  Workshops (CVPRw)}, 2020.

\bibitem[Dai et~al.(2018)Dai, Zhu, Guo, and Wipf]{2018_ICML_compressing}
Bin Dai, Chen Zhu, Baining Guo, and David Wipf.
\newblock Compressing neural networks using the variational information
  bottleneck.
\newblock In \emph{International Conference on Machine Learning (ICML)}, 2018.

\bibitem[Dean et~al.(2012)Dean, Corrado, Monga, Chen, Devin, Mao, Ranzato,
  Senior, Tucker, Yang, et~al.]{2012_NeurIPS_sgd}
Jeffrey Dean, Greg Corrado, Rajat Monga, Kai Chen, Matthieu Devin, Mark Mao,
  Marc'aurelio Ranzato, Andrew Senior, Paul Tucker, Ke~Yang, et~al.
\newblock Large scale distributed deep networks.
\newblock In \emph{Advances in Neural Information Processing Systems
  (NeurIPS)}, 2012.

\bibitem[Deng et~al.(2009)Deng, Dong, Socher, Li, Li, and
  Fei-Fei]{2009_CVPR_ImageNet}
Jia Deng, Wei Dong, Richard Socher, Li-Jia Li, Kai Li, and Li~Fei-Fei.
\newblock {ImageNet: A large-scale hierarchical image database}.
\newblock In \emph{{IEEE Conference on Computer Vision and Pattern Recognition
  (CVPR)}}, 2009.

\bibitem[Fan et~al.(2017)Fan, Lyu, Ying, and Hu]{2017_NeurIPS_topKLoss}
Yanbo Fan, Siwei Lyu, Yiming Ying, and Baogang Hu.
\newblock Learning with average top-k loss.
\newblock In \emph{Advances in Neural Information Processing systems
  (NeurIPS)}, 2017.

\bibitem[Gopal(2016)]{2016_ICML_adaptive}
Siddharth Gopal.
\newblock Adaptive sampling for sgd by exploiting side information.
\newblock In \emph{International Conference on Machine Learning}, 2016.

\bibitem[Hacohen and Weinshall(2019)]{2019_ICML_PowerOfCL}
Guy Hacohen and Daphna Weinshall.
\newblock On the power of curriculum learning in training deep networks.
\newblock In \emph{International Conference on Machine Learning (ICML)}, 2019.

\bibitem[He et~al.(2016)He, Zhang, Ren, and Sun]{2016_CVPR_ResNet}
Kaiming He, Xiangyu Zhang, Shaoqing Ren, and Jian Sun.
\newblock {Deep Residual Learning for Image Recognition}.
\newblock In \emph{{IEEE Conference on Computer Vision and Pattern Recognition
  (CVPR)}}, 2016.

\bibitem[He et~al.(2020)He, Fan, Wu, Xie, and Girshick]{CVPR_2020_moco}
Kaiming He, Haoqi Fan, Yuxin Wu, Saining Xie, and Ross Girshick.
\newblock Momentum contrast for unsupervised visual representation learning.
\newblock In \emph{IEEE Conference on Computer Vision and Pattern Recognition
  (CVPR)}, 2020.

\bibitem[Ioannou et~al.(2019)Ioannou, Tagaris, and
  Stafylopatis]{2019_ICAAI_biasedSampling}
George Ioannou, Thanos Tagaris, and Andreas Stafylopatis.
\newblock Improving the convergence speed of deep neural networks with biased
  sampling.
\newblock In \emph{International Conference on Advances in Artificial
  Intelligence (ICAAI)}, 2019.

\bibitem[Jacob et~al.(2018)Jacob, Kligys, Chen, Zhu, Tang, Howard, Adam, and
  Kalenichenko]{2018_CVPR_quantization}
Benoit Jacob, Skirmantas Kligys, Bo~Chen, Menglong Zhu, Matthew Tang, Andrew
  Howard, Hartwig Adam, and Dmitry Kalenichenko.
\newblock Quantization and training of neural networks for efficient
  integer-arithmetic-only inference.
\newblock In \emph{IEEE Conference on Computer Vision and Pattern Recognition
  (CVPR)}, 2018.

\bibitem[Jiang et~al.(2019)Jiang, Wong, Zhou, Andersen, Dean, Ganger, Joshi,
  Kaminksy, Kozuch, Lipton, et~al.]{2019_Arxiv_selective}
Angela~H Jiang, Daniel L-K Wong, Giulio Zhou, David~G Andersen, Jeffrey Dean,
  Gregory~R Ganger, Gauri Joshi, Michael Kaminksy, Michael Kozuch, Zachary~C
  Lipton, et~al.
\newblock Accelerating deep learning by focusing on the biggest losers.
\newblock \emph{arXiv:1910.00762}, 2019.

\bibitem[Jiang et~al.(2018)Jiang, Zhou, Leung, Li, and
  Fei-Fei]{2018_ICML_mentorNet}
Lu~Jiang, Zhengyuan Zhou, Thomas Leung, Li-Jia Li, and Li~Fei-Fei.
\newblock Mentornet: Learning data-driven curriculum for very deep neural
  networks on corrupted labels.
\newblock In \emph{International Conference on Machine Learning (ICML)}, pages
  2304--2313. PMLR, 2018.

\bibitem[Johnson and Guestrin(2018)]{2018_NeurIPS_RAIS}
Tyler~B Johnson and Carlos Guestrin.
\newblock Training deep models faster with robust, approximate importance
  sampling.
\newblock In \emph{Advances in Neural Information Processing Systems
  (NeurIPS)}, 2018.

\bibitem[Kachuee et~al.(2019)Kachuee, Goldstein, Karkkainen, Darabi, and
  Sarrafzadeh]{2019_ICML_opportunistic_budget}
Mohammad Kachuee, Orpaz Goldstein, Kimmo Karkkainen, Sajad Darabi, and Majid
  Sarrafzadeh.
\newblock {Opportunistic learning: Budgeted cost-sensitive learning from data
  streams}.
\newblock In \emph{{International Conference on Machine Learning (ICML)}},
  2019.

\bibitem[Katharopoulos and Fleuret(2018)]{2018_ICML_notAllSamples}
Angelos Katharopoulos and Fran{\c{c}}ois Fleuret.
\newblock Not all samples are created equal: Deep learning with importance
  sampling.
\newblock In \emph{International Conference on Machine Learning (ICML)}, 2018.

\bibitem[Kawaguchi and Lu(2020)]{2020_AISTATS_osgd}
Kenji Kawaguchi and Haihao Lu.
\newblock Ordered sgd: A new stochastic optimization framework for empirical
  risk minimization.
\newblock In \emph{International Conference on Artificial Intelligence and
  Statistics (AISTATS)}, 2020.

\bibitem[Krizhevsky et~al.(2009)Krizhevsky, Hinton, et~al.]{2009_CIFAR}
Alex Krizhevsky, Geoffrey Hinton, et~al.
\newblock {Learning multiple layers of features from tiny images}.
\newblock Technical report, University of Toronto, 2009.

\bibitem[Li et~al.(2020{\natexlab{a}})Li, Socher, and Hoi]{2020_ICLR_DivideMix}
Junnan Li, Richard Socher, and Steven~CH Hoi.
\newblock {DivideMix: Learning with Noisy Labels as Semi-supervised Learning}.
\newblock In \emph{{International Conference on Learning Representations
  (ICLR)}}, 2020{\natexlab{a}}.

\bibitem[Li et~al.(2020{\natexlab{b}})Li, Yumer, and Ramanan]{2020_ICLR_budget}
Mengtian Li, Ersin Yumer, and Deva Ramanan.
\newblock {Budgeted training: Rethinking deep neural network training under
  resource constraints}.
\newblock In \emph{{International Conference on Learning Representations
  (ICLR)}}, 2020{\natexlab{b}}.

\bibitem[Lin et~al.(2019)Lin, Ji, Yan, Zhang, Cao, Ye, Huang, and
  Doermann]{2019_CVPR_compressGAN}
Shaohui Lin, Rongrong Ji, Chenqian Yan, Baochang Zhang, Liujuan Cao, Qixiang
  Ye, Feiyue Huang, and David Doermann.
\newblock Towards optimal structured cnn pruning via generative adversarial
  learning.
\newblock In \emph{IEEE Conference on Computer Vision and Pattern Recognition
  (CVPR)}, 2019.

\bibitem[Lin et~al.(2017)Lin, Goyal, Girshick, He, and
  Doll{\'a}r]{2017_ICCV_focal_loss}
Tsung-Yi Lin, Priya Goyal, Ross Girshick, Kaiming He, and Piotr Doll{\'a}r.
\newblock Focal loss for dense object detection.
\newblock In \emph{International Conference on Computer Vision (ICCV)}, pages
  2980--2988, 2017.

\bibitem[Loshchilov and Hutter(2015)]{2015_ICLRw_online}
Ilya Loshchilov and Frank Hutter.
\newblock Online batch selection for faster training of neural networks.
\newblock \emph{arXiv:1511.06343}, 2015.

\bibitem[Lu and Mazumder(2018)]{2018_Arxiv_randGradBoost}
Haihao Lu and Rahul Mazumder.
\newblock Randomized gradient boosting machine.
\newblock \emph{arXiv:1810.10158}, 2018.

\bibitem[Mahajan et~al.(2018)Mahajan, Girshick, Ramanathan, He, Paluri, Li,
  Bharambe, and van~der Maaten]{2018_ECCV_limits_weakly}
Dhruv Mahajan, Ross Girshick, Vignesh Ramanathan, Kaiming He, Manohar Paluri,
  Yixuan Li, Ashwin Bharambe, and Laurens van~der Maaten.
\newblock Exploring the limits of weakly supervised pretraining.
\newblock In \emph{European Conference on Computer Vision (ECCV)}, September
  2018.

\bibitem[Mirzasoleiman et~al.(2020)Mirzasoleiman, Bilmes, and
  Leskovec]{2020_ICML_coreSet}
Baharan Mirzasoleiman, Jeff Bilmes, and Jure Leskovec.
\newblock Coresets for data-efficient training of machine learning models.
\newblock In \emph{International Conference on Machine Learning (ICML)}, 2020.

\bibitem[Misra and Maaten(2020)]{CVPR_2020_InvRepresentations}
Ishan Misra and Laurens van~der Maaten.
\newblock Self-supervised learning of pretext-invariant representations.
\newblock In \emph{IEEE Conference on Computer Vision and Pattern Recognition
  (CVPR)}, 2020.

\bibitem[Nan and Saligrama(2017)]{2017_NeurIPS_adaptive_budget}
Feng Nan and Venkatesh Saligrama.
\newblock Adaptive classification for prediction under a budget.
\newblock In \emph{Advances in neural information processing systems
  (NeurIPS)}, 2017.

\bibitem[Needell et~al.(2014)Needell, Ward, and Srebro]{2014_NeurIPS_sgdweight}
Deanna Needell, Rachel Ward, and Nati Srebro.
\newblock Stochastic gradient descent, weighted sampling, and the randomized
  kaczmarz algorithm.
\newblock In \emph{Advances in neural information processing systems
  (NeurIPS)}, 2014.

\bibitem[Netzer et~al.(2011)Netzer, Wang, Coates, Bissacco, Wu, and
  Ng]{2011_NeurIPS_SVHN}
Y.~Netzer, T.~Wang, A.~Coates, A.~Bissacco, B.~Wu, and A.Y. Ng.
\newblock Reading digits in natural images with unsupervised feature learning.
\newblock In \emph{{Advances in Neural Information Processing Systems
  (NeurIPS)}}, 2011.

\bibitem[Rastegari et~al.(2016)Rastegari, Ordonez, Redmon, and
  Farhadi]{2016_ECCV_xnor}
Mohammad Rastegari, Vicente Ordonez, Joseph Redmon, and Ali Farhadi.
\newblock Xnor-net: Imagenet classification using binary convolutional neural
  networks.
\newblock In \emph{European Conference on Computer Vision (ECCV)}, 2016.

\bibitem[Ren et~al.(2020)Ren, Xiao, Chang, Huang, Li, Chen, and
  Wang]{2020_ArXiv_active_learning_survey}
Pengzhen Ren, Yun Xiao, Xiaojun Chang, Po-Yao Huang, Zhihui Li, Xiaojiang Chen,
  and Xin Wang.
\newblock A survey of deep active learning.
\newblock \emph{arXiv: 2009.00236}, 2020.

\bibitem[Sener and Savarese(2018)]{2018_ICLR_activelearning}
Ozan Sener and Silvio Savarese.
\newblock {Active learning for convolutional neural networks: A core-set
  approach}.
\newblock In \emph{{International Conference on Learning Representations
  (ICLR)}}, 2018.

\bibitem[Smith(2017)]{2017_WACV_cyclical}
Leslie~N Smith.
\newblock Cyclical learning rates for training neural networks.
\newblock In \emph{IEEE Winter Conference on Applications of Computer Vision
  (WACV)}, 2017.

\bibitem[Smith and Topin(2019)]{2019_ISOP_superConv}
Leslie~N Smith and Nicholay Topin.
\newblock Super-convergence: Very fast training of neural networks using large
  learning rates.
\newblock In \emph{Artificial Intelligence and Machine Learning for
  Multi-Domain Operations Applications}, 2019.

\bibitem[Sun et~al.(2019)Sun, Xiao, Liu, and Wang]{2019_CVPR_pose_Estimation}
Ke~Sun, Bin Xiao, Dong Liu, and Jingdong Wang.
\newblock Deep high-resolution representation learning for human pose
  estimation.
\newblock In \emph{IEEE Conference on Computer Vision and Pattern Recognition
  (CVPR)}, 2019.

\bibitem[Takahashi et~al.(2018)Takahashi, Matsubara, and
  Uehara]{2018_ACML_RICAP}
Ryo Takahashi, Takashi Matsubara, and Kuniaki Uehara.
\newblock Ricap: Random image cropping and patching data augmentation for deep
  cnns.
\newblock In \emph{Asian Conference on Machine Learning (ACML)}, 2018.

\bibitem[Tan and Le(2019)]{2019_ICML_efficientnet}
Mingxing Tan and Quoc~V Le.
\newblock {Efficientnet: Rethinking model scaling for convolutional neural
  networks}.
\newblock In \emph{{International Conference on Machine Learning (ICML)}},
  2019.

\bibitem[Toneva et~al.(2018)Toneva, Sordoni, Combes, Trischler, Bengio, and
  Gordon]{2018_ICLR_forget}
Mariya Toneva, Alessandro Sordoni, Remi Tachet~des Combes, Adam Trischler,
  Yoshua Bengio, and Geoffrey~J Gordon.
\newblock {An empirical study of example forgetting during deep neural network
  learning}.
\newblock In \emph{{International Conference on Learning Representations
  (ICLR)}}, 2018.

\bibitem[Touvron et~al.(2019)Touvron, Vedaldi, Douze, and
  J{\'e}gou]{2019_NeurIPS_fixing}
Hugo Touvron, Andrea Vedaldi, Matthijs Douze, and Herv{\'e} J{\'e}gou.
\newblock Fixing the train-test resolution discrepancy.
\newblock In \emph{Advances in Neural Information Processing Systems
  (NeurIPS)}, 2019.

\bibitem[Vinyals et~al.(2016)Vinyals, Blundell, Lillicrap, Kavukcuoglu, and
  Wierstra]{2016_NIPS_MiniImageNet}
O.~Vinyals, C.~Blundell, T.~Lillicrap, K.~Kavukcuoglu, and D.~Wierstra.
\newblock {Matching Networks for One Shot Learning}.
\newblock In \emph{{Advances in Neural Information Processing Systems
  (NeurIPS)}}, 2016.

\bibitem[Weinshall et~al.(2018)Weinshall, Cohen, and
  Amir]{2018_ICML_CLTransfLearn}
Daphna Weinshall, Gad Cohen, and Dan Amir.
\newblock Curriculum learning by transfer learning: Theory and experiments with
  deep networks.
\newblock In \emph{International Conference on Machine Learning (ICML)}, 2018.

\bibitem[Xie et~al.(2020)Xie, Luong, Hovy, and Le]{2020_CVPR_improveImagenet}
Qizhe Xie, Minh-Thang Luong, Eduard Hovy, and Quoc~V. Le.
\newblock Self-training with noisy student improves imagenet classification.
\newblock In \emph{IEEE Conference on Computer Vision and Pattern Recognition
  (CVPR)}, June 2020.

\bibitem[Yao et~al.(2018)Yao, Pan, Li, and Mei]{2018_ECCV_captioning}
Ting Yao, Yingwei Pan, Yehao Li, and Tao Mei.
\newblock Exploring visual relationship for image captioning.
\newblock In \emph{European Conference on Computer Vision (ECCV)}, 2018.

\bibitem[Yoo and Kweon(2019)]{2019_CVPR_learning_loss_for_active_learning}
Donggeun Yoo and In~So Kweon.
\newblock Learning loss for active learning.
\newblock In \emph{IEEE Conference on Computer Vision and Pattern Recognition
  (CVPR)}, 2019.

\bibitem[Zhang et~al.(2019{\natexlab{a}})Zhang, {\"O}ztireli, Mandt, and
  Salvi]{2019_AAAI_repulsive}
Cheng Zhang, Cengiz {\"O}ztireli, Stephan Mandt, and Giampiero Salvi.
\newblock Active mini-batch sampling using repulsive point processes.
\newblock In \emph{AAAI Conference on Artificial Intelligence},
  2019{\natexlab{a}}.

\bibitem[Zhang et~al.(2018)Zhang, Cisse, Dauphin, and
  Lopez-Paz]{2018_ICLR_mixup}
Hongyi Zhang, Moustapha Cisse, Yann~N Dauphin, and David Lopez-Paz.
\newblock {mixup: Beyond empirical risk minimization}.
\newblock In \emph{{International Conference on Learning Representations
  (ICLR)}}, 2018.

\bibitem[Zhang et~al.(2019{\natexlab{b}})Zhang, Yu, and
  Dhillon]{2019_NeurIPS_autoAssist}
Jiong Zhang, Hsiang-Fu Yu, and Inderjit~S Dhillon.
\newblock Autoassist: A framework to accelerate training of deep neural
  networks.
\newblock In \emph{Advances in Neural Information Processing Systems
  (NeurIPS)}, 2019{\natexlab{b}}.

\bibitem[Zhao and Zhang(2015)]{2015_ICML_stochasticIS}
Peilin Zhao and Tong Zhang.
\newblock Stochastic optimization with importance sampling for regularized loss
  minimization.
\newblock In \emph{International Conference on Machine Learning (ICML)}, 2015.

\bibitem[Zhou et~al.(2020)Zhou, Cui, Jia, Yang, and
  Tian]{2020_CVPR_fewshot_base}
Linjun Zhou, Peng Cui, Xu~Jia, Shiqiang Yang, and Qi~Tian.
\newblock Learning to select base classes for few-shot classification.
\newblock In \emph{IEEE Conference on Computer Vision and Pattern Recognition
  (CVPR)}, 2020.

\end{thebibliography}
\end{document}